\documentclass[letterpaper,twocolumn,10pt]{article}  

\usepackage[utf8]{inputenc}
\usepackage{times}
\usepackage{graphicx}
\usepackage[left=0.75in,right=0.75in,top=1in,bottom=1in]{geometry}
\usepackage{booktabs}
\usepackage{amsmath,amssymb}
\usepackage{hyperref}
\usepackage{microtype}
\usepackage{enumitem}
\usepackage{caption}
\usepackage{subcaption}
\usepackage{balance}
\usepackage{algorithm}
\usepackage{algpseudocode}
\usepackage{forest}
\usepackage{tikz}
\usetikzlibrary{arrows.meta, positioning}
\makeatletter
\renewenvironment{thebibliography}[1]
     {\section*{\refname}%
      \list{\@biblabel{\arabic{enumiv}}}%
           {\settowidth\labelwidth{\@biblabel{#1}}%
            \leftmargin\labelwidth
            \advance\leftmargin\labelsep
            \usecounter{enumiv}%
            \setlength{\itemsep}{0pt}
            \setlength{\parsep}{0pt}
           }%
      \sloppy\clubpenalty4000\widowpenalty4000%
      \sfcode`\.=\@m}
     {\endlist}
\makeatother

\title{Adversarial Reinforcement Learning for Offensive and Defensive Agents in a Simulated Zero-Sum Network Environment}
\author{
    Abrar Shahid$^1$, Ibteeker Mahir Ishum$^1$, AKM Tahmidul Haque$^1$, \\
    M Sohel Rahman$^2$, A. B. M. Alim Al Islam Razi$^2$ \\[0.5em]
    $^1$ Notre Dame College, Dhaka \\
    $^2$ Department of Computer Science and Engineering, \\
    Bangladesh University of Engineering and Technology \\
}
\date{}

\begin{document}
\maketitle

\begin{abstract}
We present a controlled study of two competing reinforcement learning agents in a custom OpenAI Gym-style environment that models offensive brute-force attacks and reactive defenses on a multi-port service. The environment captures realistic trade-offs that model background traffic, brute-force exploits, IP-based evasion, traps, and rate-limiting defenses. Agents are trained using deep Q networks (DQNs) with a zero-sum reward structure. Successful exploits give large terminal rewards, while step actions incur small costs. We evaluated value-based agents in multiple locations, including trap probability, exploit difficulty, and training regimen. The results demonstrate that the observability of the defender and the effectiveness of the trap strongly hinder exploitations. In this scenario, reward shaping and training scheduling are crucial for learning stability. We provide implementation details, reproducible configurations, and guidance for future extensions.
\end{abstract}

\section{Introduction}
In recent years, cyberattacks have become increasingly sophisticated that cause widespread damage to critical infrastructure, financial systems, and everyday online services~\cite{globalcyber}. High-profile incidents such as large-scale ransomware outbreaks and coordinated brute-force attacks demonstrate how traditional rule-based defenses often fail to adapt quickly enough. This motivates the need for autonomous systems that can proactively respond to evolving threats. 

Automated decision systems increasingly mediate interactions between attackers and defenders in networked systems~\cite{automatedsystem}. Research on such interactions benefits from controlled simulation environments that allow repeatable experiments and careful measurement of algorithmic behaviors. We introduce a Gym environment focusing on offensive brute-force-type exploitation and on defensive countermeasures that include IP rate-limiting, port rate-limiting, traps, and port closure.~\cite{gymnasium2024} ~\cite{openaigym2016} The environment reflects two important properties: \\ (1) Normal user traffic produces substantial noise obscuring attacks, and  \\ (2) Successful exploitation requires sustained request volume, creating a tradeoff for the attacker between persistence and detectability.

\subsection{Reinforcement Learning (RL)}
\textbf{Reinforcement Learning (RL)} provides a framework for agents to learn optimal strategies through interaction with an environment. At each discrete timestep, an agent observes a state $s \in \mathcal{S}$, selects an action $a \in \mathcal{A}$ according to a policy $\pi(a|s)$, and receives a reward $r \in \mathbb{R}$ that quantifies the immediate outcome, where $\mathcal{S}$ refers to the observation space, $\mathcal{A}$ refers to the action space, and $\mathbb{R}$ refers to the rewards possible under discrete actions. The agent's goal is to learn a policy that maximizes the expected cumulative reward over time.~\cite{rlsystems} \\ Value-based methods, such as Q-learning~\cite{qlearning} and Deep Q-Networks (DQN) ~\cite{dqn_thesis2022} , estimate the action-value function $Q(s,a)$ to guide decision-making. Exploration strategies like $\varepsilon$-greedy are typically used to balance trying new actions with exploiting known good actions.
\vspace{1cm} 
\begin{figure}[h]
    \centering
    \begin{tikzpicture}[
        ->, >=Stealth, auto,
        block/.style={rectangle, draw, rounded corners, minimum height=2em, minimum width=3cm, align=center},
        node distance=2cm
    ]

    \node[block] (agent) {Agent};
    \node[block, below=of agent] (env) {Environment};

    \draw[->] (agent) -- node[right] {$a_t$} (env);
    \draw[->] (env) -- node[left] {$r_t$} (agent);

    \draw[->] (env.west) -- ++(-3,0) |- node[pos=0.25, left] {$s_{t+1}$} (agent.west);
    \draw[->] (agent.west) -- ++(-3,0) |- node[pos=0.75, left] {$s_t$} (env.west);

    \end{tikzpicture}
    \caption{Reinforcement Learning agent workflow in an environment}
    \label{fig:rl_agent}
\end{figure}
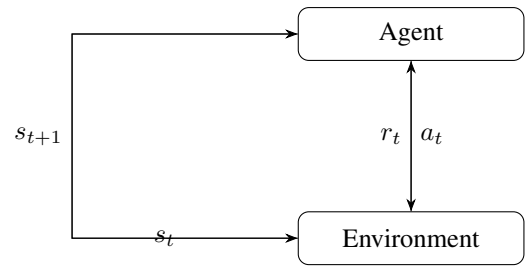

\subsection{Markov Decision Process (MDP)} 
A Markov Decision Process (MDP) ~\cite{mdp} is a mathematical framework commonly used in reinforcement learning to model decision-making problems. An MDP is defined as the tuple $\langle \mathcal{S}, \mathcal{A}, \mathcal{P}, \mathcal{R} \rangle$, where:
\begin{itemize}[noitemsep]
    \item $\mathcal{S}$ is the finite set of states,
    \item $\mathcal{A}$ is the finite set of actions available in each state,
    \item $\mathcal{P} : \mathcal{S} \times \mathcal{A} \rightarrow \mathcal{S}$ is the state transition function, and
    \item $\mathcal{R} : \mathcal{S} \times \mathcal{A} \rightarrow \mathbb{R}$ is the immediate reward function.
\end{itemize}

In our zero-sum cyber security simulation, the attacker and defender interact sequentially through this MDP. At each timestep $t$, the attacker performs action $a_t \in \mathcal{A}(s_t)$ on state $s_t \in \mathcal{S}$, producing an intermediate state $s_t' = \psi(s_t, a_t)$ (e.g., a successful scan or exploit attempt). The environment responds with a reward $r_t = \mathcal{R}(s_t, a_t)$ and updates to the next state $s_{t+1} \sim \mathcal{P}(s_t', s_{t+1})$ (e.g., defender response such as rate-limiting or trap activation). The process repeats until a terminal state $s_T \in \mathcal{S}$ is reached, where the attacker or defender cannot act further.

The goal in this zero-sum environment is to find a policy $\pi$ for each agent that maximizes its cumulative reward while minimizing the opponent’s gain. The state-value function for the attacker, for instance, is defined as
\[
V(s_t) = \mathbb{E}[r_t + r_{t+1} + \dots],
\]
\begin{flushright} \dots\dots\dots (1) \end{flushright}
and the optimal policy can be derived as
\[
\pi(s_t) = \arg\max_{a_t \in \mathcal{A}(s_t)} \Big( r_t + \sum_{s_{t+1}} \mathcal{P}(s_t', s_{t+1}) V(s_{t+1}) \Big),
\]
\begin{flushright} \dots\dots\dots (2) \end{flushright}
Here $s_t' = \psi(s_t, a_t)$ represents the intermediate state after the action.

This paper reports a set of experiments that use value-based agents to study learning dynamics, policy stability, and strategic outcomes. We aim to identify the factors that favor defensive or offensive success and to provide methodological recommendations for further work. Our contributions are: \vspace{0.5em}\\ (1) A detailed environment design that captures brute-force exploits and defender actions; \\ (2) An empirical study across multiple ablations; and \\ (3) Practical recommendations to improve training stability and interoperability.

\vspace{0.5em}

\section{Methodology}

This section describes the environment, the agents, and the experimental protocol. We give precise definitions of actions, observations, rewards, and the exploitation mechanics.

\subsection{Environment Overview}
The environment simulates a multi-agent zero-sum cyber security scenario, modeling a host with $N$ independent ports. In each simulation there are $M$ IPs among which most are normal users and some are reserved for attacker which are randomized on each episode. At the beginning of each episode, a random subset of ports is designated as \emph{vulnerable}, and each vulnerable port is assigned an exploitation threshold $T_p$, sampled randomly with a minimum value $T_{\min}=300$. Exploitation progresses by sending attack-specific requests to the target port from a source IP address. The attacker can change its IP periodically, simulating the use of proxies in real-world attacks, but only after a predefined number of actions.

The environment also generates background normal traffic at each timestep to represent legitimate network activity. Normal traffic is generated from a pool of IP addresses, and the defender must avoid blocking legitimate users while mitigating attacks.  

Time evolves in discrete steps, and at each step, only one agent acts: first the attacker, then the defender. The defender is given the option to implement no action in certain steps and rather observe the outcomes. Some actions have instantaneous effects (e.g., scanning a port, setting a trap, or closing a port), while others initiate continuous processes (e.g., an exploit attempt accumulates requests over multiple timesteps). The environment enforces action alternation between agents, but continuous processes such as exploit progression and normal traffic generation proceed independently between actions.

The environment maintains a history of requests, including attacker and normal traffic, to calculate observations for both agents. Observations for the attacker include scanned ports, exploit progress, and current IP status, while defender observations include port request counts, suspicious activity indicators, top IP activity, and current defense configurations. Rewards are structured to enforce a zero-sum dynamic: attacker gains correspond to defender losses and vice versa.

\subsection{Action Spaces}
In reinforcement learning, the \emph{action space} $\mathcal{A}$ defines the set of all possible actions an agent can take at a given timestep. Each action $a \in \mathcal{A}$ influences the environment state and contributes to the cumulative reward objective. In our zero-sum adversarial setting, both the attacker and defender are restricted to discrete, finite action spaces to ensure tractability and reproducibility. 

They are limited to choose each action from the action spaces depending on the scenario.

\begin{itemize}[noitemsep]
\item \textbf{Attacker}: Scan port $i$ for $i\in\{1\ldots N\}$; Exploit port $i$ for $i\in\{1\ldots N\}$; Change IP (after a minimum of 10 actions) ; Cancel exploit.
\\ \\
The attacker can: Scan ports to detect vulnerabilities, Launch exploits against discovered vulnerable ports, which accumulate over successive timesteps until success, failure, or interruption, Change its IP address to evade detection and bypass rate limits.Cancel ongoing exploits to adjust strategy in response to defender actions.
\\ \\
\item \textbf{Defender}: Wait(No action); Rate-limit IP $q$ for $q \in \{1, \dots, M\}$; Rate-limit port $j$ for $j\in\{1\ldots N\}$; Set trap on port $k$ for $k\in\{1\ldots N\}$; Close port $l$ for $l\in\{1\ldots N\}$
\\ \\
The defender can: Rate-limit traffic from specific IP addresses or ports, Rate-limit an IP to a number of requests, Deploy traps on ports to penalize attackers, Close ports to completely block access or Simply initiate no action for that step and rather observe the outcomes for further strategies. The ports here are discrete.

\end{itemize}

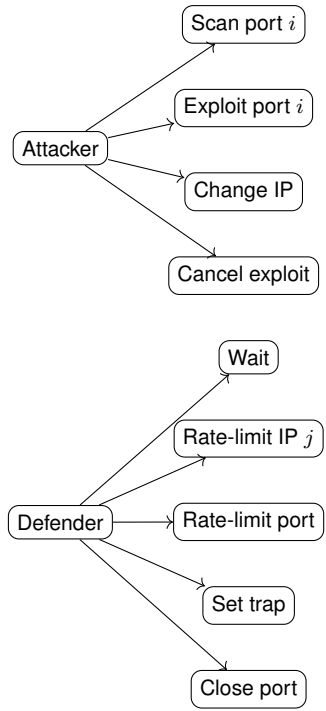
\begin{figure}[h]
\centering
\begin{forest}
for tree={
  draw, rounded corners,
  node options={align=center, font=\sffamily\footnotesize},
  grow'=0, 
  edge={->},
  s sep=6mm, 
  l sep=8mm  
}
[,phantom 
  [Attacker
    [Scan port $i$]
    [Exploit port $i$]
    [Change IP]
    [Cancel exploit]
  ]
  [Defender
    [Wait]
    [Rate-limit IP $j$]
    [Rate-limit port]
    [Set trap]
    [Close port]
  ]
]
\end{forest}
\caption{Flowchart of attacker and defender action spaces.}
\end{figure}

These discrete actions capture standard reconnaissance, exploitation, and defense primitives while keeping the action dimensions tractable.

\subsection{Observation Spaces}
The \emph{observation space} $\mathcal{O}$ defines the set of all possible observations an agent can receive from the environment at each timestep. Each observation $o \in \mathcal{O}$ encodes relevant information about the environment state that the agent can use to select actions and maximize its expected cumulative reward. Each continuous observation is discretized into 3 bins, striking a balance between state variability and memory usage. 
Formally, the state vector for an agent at timestep $t$ is $s_t \in \mathcal{S} \subset \mathbb{R}^{d_{\text{obs}}}$, which is normalized and fed into the Deep Q-Network. This representation allows the agents to learn policies despite partial observability of the environment.

\textbf{Attacker:} The attacker receives a continuous observation vector encoding the results of recent port scans, exploit progress indicators, and a short history of its past actions with their outcomes. Scan results capture both whether a port appears vulnerable and whether defensive anomalies were triggered. An anomaly signal suggests a probabilistic trap by the defender. The attacker also observes its current IP status (e.g., active, or blacklisted).

\textbf{Defender:} The defender receives a continuous observation vector summarizing traffic statistics per port, suspicious activity ratios aggregated over source IPs, and a sliding window of attacker IP histories. In addition, it maintains awareness of the current status of defenses (rate limits, traps, or port closures) and their effectiveness in past interactions. This reflects the log-based perspective of a real-world security operations center, where defenders rely on accumulated traffic data and historical patterns to anticipate intrusions using global tools like SIEM and Elastic.

\subsection{Exploration Strategy}
The exploration strategy for both the attacker and defender agents is based on the epsilon-greedy approach~\cite{epsilon_greedy_study}. In the early training phases, the agents are encouraged to explore the environment by selecting random actions with high probability. As training progresses, the exploration rate decays, and the agents gradually shift towards exploitation of learned strategies. 

\[
a_t = \left\{
\begin{array}{ll}
\arg\max Q(s_t, a) & \text{with probability } 1 - \varepsilon, \\
\text{random action from } \mathcal{A} & \text{with probability } \varepsilon
\end{array}
\right\}
\]
\begin{flushright} \dots\dots\dots (3) \end{flushright}
\begin{flushleft}
Here $a_t$ is the action taken, $\mathcal{A}$ is the action space and $\arg\max Q(s_t, a)$ is the best learned action.
\end{flushleft}

The table below outlines the exploration parameters for both agents:

\begin{table}[ht]
\centering
\caption{Exploration Parameters for the Attacker Agent}
\begin{tabular}{|c|c|}
\hline
\textbf{Parameter} & \textbf{Value} \\ \hline
Learning Rate ($\alpha$) & 0.001 \\ \hline
Discount Factor ($\gamma$) & 0.95 \\ \hline
Initial Exploration ($\epsilon$) & 1.0 \\ \hline
Epsilon Decay Rate & 0.995 \\ \hline
Minimum Epsilon ($\epsilon_{\min}$) & 0.05 \\ \hline
\end{tabular}
\label{tab:attacker_params}
\end{table}

\begin{table}[ht]
\centering
\caption{Exploration Parameters for the Defender Agent}
\begin{tabular}{|c|c|}
\hline
\textbf{Parameter} & \textbf{Value} \\ \hline
Learning Rate ($\alpha$) & 0.002 \\ \hline
Discount Factor ($\gamma$) & 0.90 \\ \hline
Initial Exploration ($\epsilon$) & 1.0 \\ \hline
Epsilon Decay Rate & 0.99 \\ \hline
Minimum Epsilon ($\epsilon_{\min}$) & 0.05 \\ \hline
Batch Size & 512 \\ \hline
\end{tabular}
\label{tab:defender_params}
\end{table}

These parameters ensure both agents to learn efficiently from the environment. The learning rate (\(\alpha\)) determines how quickly the agents update their policies, while the discount factor (\(\gamma\)) ensures the balance of long-term rewards with short-term gains. ~\cite{hyperparameters_rl}. For the attacker, a rate of 0.001 facilitates faster convergence to an optimal policy, while for the defender, a slightly higher rate of 0.002 enables quicker adaptation to the attacker's strategies. Both agents use a high discount factor to balance immediate and future rewards. Initially, both agents explore the environment aggressively, but over time the exploration rate decays, helping the agents shift from exploration to exploitation as they learn optimal strategies.
For the attacker, the decay rate is 0.995, meaning it reduces exploration slightly slower than the defender, whose decay rate is 0.99.
These rates enable the agents to explore aggressively in the early stages, while reducing exploration as they approach optimal policies.
The epsilon decay along each episode is as follows:
\begin{gather*}
\varepsilon(t) = \max\big(\varepsilon_{\min}, \varepsilon_{\text{initial}} \cdot (\text{decay\_rate})^t\big), \\
\text{When } \varepsilon(t) = \varepsilon_{\min}, \quad
t = \frac{\ln\left(\frac{\varepsilon_{\min}}{\varepsilon_{\text{initial}}}\right)}{\ln(\text{decay\_rate})}
\end{gather*}

\begin{flushright} \dots\dots\dots (4) \end{flushright}
Nearly after t episodes with progressive epsilon decay, the defensive or offensive algorithm follows the best strategy. The epsilon hits the minimum value resulting in the algorithm to always choose the action $\arg\max Q(s_t, a)$ with highest probability.

\subsection{Episode Mechanism}
\subsubsection{Exploitation Mechanics}
Exploitation is modeled as a threshold accumulation of attack requests per (IP, port) pair. Each exploit attempt increments a counter $c_{ip,p}$. An exploit succeeds if $c_{ip,p}\ge T_p$ for a vulnerable port $p$. The threshold $T_p$ is sampled at episode start with $T_p\ge T_{min}$. The attacker may change its source IP after at least 10 requests from the current IP; changing the IP resets the attacker-specific counter for the next IP. So, The attacker needs to be careful regarding his choice for IP change since changing the IP resets the counter while as detection of multiple request from same IP might result in rate limiting

\subsubsection{Defense Mechanics}
A trap placed on a port has a probability $P_{detect}$ of being indicated as an anomaly when scanned, and penalizes attackers who reach the trap while exploiting. Traps are intended to create a strategic deterrent and increase defender rewards when triggered.
Rate limiting a port to request capacity hampers benign user requests as well. Closing a port implies shutting down a service to all users and incurring a great loss to defend the exploit.

\section{Rewards Design}
Crucially, the interaction between attacker and defender is inherently a \textit{zero-sum environment}: every successful exploitation by the attacker represents a direct loss for the defender, while effective defense simultaneously denies the attacker’s reward.~\cite{zerosumgames} This adversarial framing makes RL particularly well-suited for studying cyber conflict, as it mirrors the strategic interplay where one side’s gain is exactly the other’s harm. ~\cite{reward_design}.
\[
r_{\text{attacker}} = -\,r_{\text{defender}}.
\]

\begin{table}[ht]
\centering
\caption{Attacker Reward Design}
\begin{tabular}{|c|c|}
\hline
\textbf{Action} & \textbf{Reward} \\ \hline
Successful Exploit & +100 \\ \hline
Trap Hit & -80 \\ \hline
Scan Cost & -0.125 \\ \hline
Exploit Attempt & -0.25 \\ \hline
Cancel Exploit & -4 \\ \hline
Change IP cost & -8 \\ \hline
\end{tabular}
\label{tab:general_attacker_rewards}
\end{table}

\begin{table}[ht]
\centering
\caption{Defender Reward Design}
\begin{tabular}{|c|c|}
\hline
\textbf{Action} & \textbf{Reward} \\ \hline
Successful Defense & 100 \\ \hline
Rate limit IP cost & -8 \\ \hline
Rate limit Port Cost & -12 \\ \hline
Close port cost & -40 \\ \hline
Trap Set & -4 \\ \hline
Block normal request & -8 \\ \hline
\end{tabular}
\label{tab:general_defender_rewards}
\end{table}

\begin{itemize}
  \item \textbf{Attacker rewards and cost}: Table~\ref{tab:general_attacker_rewards} shows the detailed reward structure for the attacker. The attacker gains a significant reward for successfully exploiting a vulnerability and faces a penalty when falling into traps or failing exploits. Probing actions incur a small cost but can provide valuable information about vulnerabilities or honeypots.

\item \textbf{Defender rewards and cost}:
The defender's rewards are linked to its defensive actions, such as setting traps or rate-limiting the attacker. Table ~\ref{tab:general_defender_rewards} summarizes the defender's reward design.
The defender earns rewards for successfully blocking attacks, such as by trapping the attacker or rate-limiting its actions. However, penalties are imposed for defensive mistakes like blocking legitimate traffic or unnecessarily closing ports. ~\cite{reward_shaping_grzes}
\end{itemize}

\section{Agent Training and Evaluation}
We trained the agents in 2 environments enabling simple to complex strategies in the end. Each environment had the same technical scope but trained on different algorithms, parameters and scope for coming up with complex strategies.

\subsection{Model Structure}
In our model, we employ a Deep Q-Network for both the attacker and defender agents. This approach allows both agents to adapt their strategies based on the rewards received from their actions.~\cite{dqnl} Both the attacker and defender use a sparse Q-table implementation, represented as a dictionary. This structure significantly reduces memory usage, which is crucial for handling large state spaces efficiently. Both agents update their Q-values after each action using the Q-learning update rule:
\vspace{0.5cm}
{\footnotesize{ \[ Q(s_t, a_t) \leftarrow Q(s_t, a_t)  + \alpha \left[ r_t + \gamma \max_{a'} Q(s_{t+1}, a') - Q(s_t, a_t) \right] \]}} \begin{flushright} \dots\dots\dots (5) \end{flushright} 
Here \(Q(s_t, a_t)\) is the Q-value for the state-action pair \((s_t, a_t)\), \(r_t\) is the reward at time step \(t\), and \(\gamma\) is the discount factor.
The DQN maps the agent's observation vector to Q-values for all possible discrete actions:
\[ Q: \mathbb{R}^{d_{\text{obs}}} \rightarrow \mathbb{R}^{d_{\text{action}}}, \quad Q(s) = [Q(s,a_1), \dots, Q(s,a_{d_{\text{action}}})],\]
\begin{flushright} \dots\dots\dots (6) \end{flushright}
Here,
\begin{itemize}[noitemsep]
    \item $d_{\text{obs}}$ is the dimension of the observation vector 
    \item $d_{\text{action}}$ is the number of discrete actions 
    \item $s \in \mathbb{R}^{d_{\text{obs}}}$ is the current observation
    \item $Q(s,a)$ estimates the expected cumulative reward for taking action $a$ in state $s$
\end{itemize}
\begin{algorithm}
\caption{Deep Q-Network (DQN) with $\varepsilon$-Greedy Policy}
\begin{algorithmic}[1]
\State Initialize replay memory $\mathcal{D}$
\State Initialize Q-network with random weights $\theta$
\State Initialize target network $\theta^- \gets \theta$
\For{each episode}
    \State Initialize state $s_0$
    \For{each step $t$}
        \State Choose action $a_t$ using $\varepsilon$-greedy from $Q(s_t;\theta)$
        \State Execute $a_t$, observe reward $r_{t+1}$ and next state $s_{t+1}$
        \State Store $(s_t, a_t, r_{t+1}, s_{t+1})$ in $\mathcal{D}$
        \If{enough samples in $\mathcal{D}$}
            \State Sample random minibatch from $\mathcal{D}$
            \State Compute target: $y = r + \gamma \max_{a'} Q(s', a'; \theta^-)$
            \State Update $\theta$ to minimize $(y - Q(s,a;\theta))^2$
        \EndIf
        \State Periodically update target network: $\theta^- \gets \theta$
        \State $s_t \gets s_{t+1}$
    \EndFor
\EndFor
\end{algorithmic}
\end{algorithm}

\noindent Here the algorithm mainly represents how the model optimizes for best action after each timesteps with a balance of exploration and exploitation.

\begin{figure*}[h]
    \centering
    \includegraphics[width=0.7\linewidth]{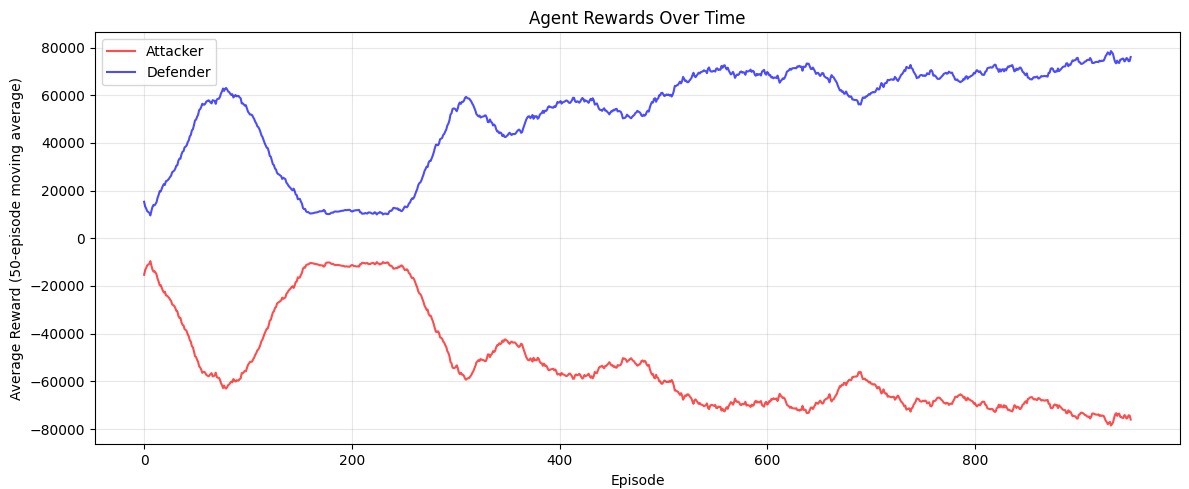}
    \caption{Early training dynamics with 50-episode moving average rewards. The defender achieves rapid positive escalation while attacker rewards remain deeply negative, illustrating strong divergence in the initial learning phase.}
    \label{fig:initial_reward_dynamics}
\end{figure*}

\begin{figure*}
    \centering
    \includegraphics[width=0.85\linewidth]{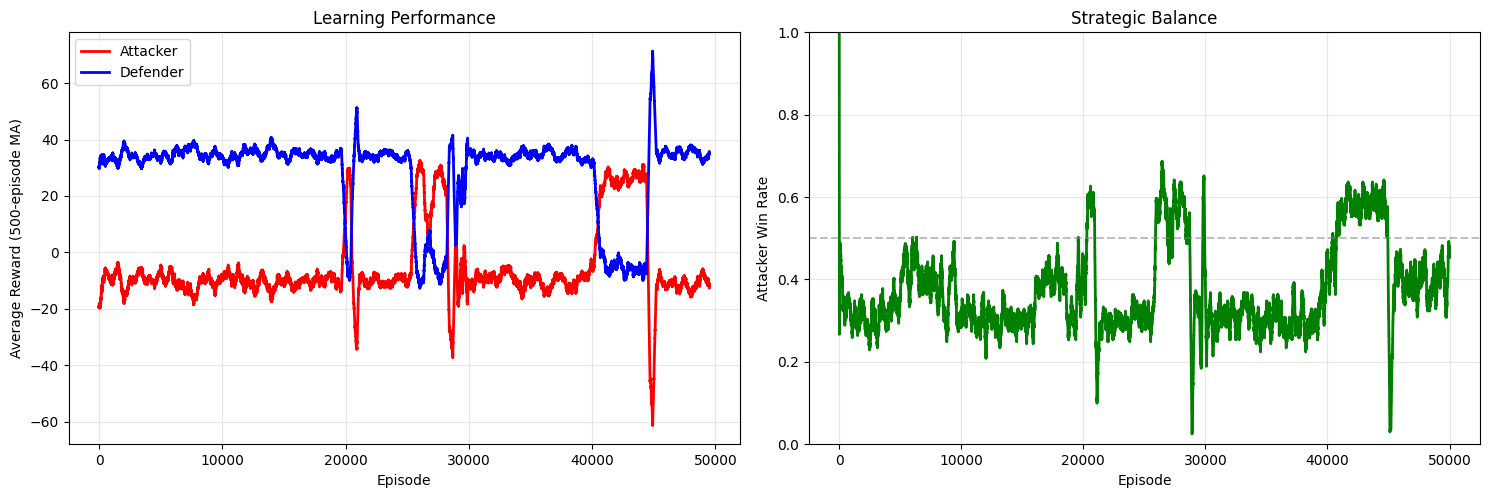}  
    \caption{Learning Performance of Attacker and Defender Agents during initial training}
    \label{fig:model1learning_rate}
\end{figure*}

\subsection{Parameters Used}
\begin{table}[H]   
\centering
\caption{Primary hyperparameter (representative).}
\begin{tabular}{lrr}
\toprule
Hyperparameter & Value \\
\midrule
Episodes & 20000 \\
Batch size & 512 \\
Replay buffer & 75{,}000 \\
$T_{min}$ & 300 req \\
Normal requests per action & 50-70 \\
Ports Used $N$ & 10-15\\
Vulnerable ports & 3-7\\
Trap detection & 60\% \\
Max previous history states & 150 \\
\bottomrule
\end{tabular}
\label{tab:hyper}
\end{table}

The large batch size of 512 samples leverages GPU parallelization for efficient neural network updates while providing stable gradient estimates that reduce learning variance. The substantial replay buffer capacity of 75,000 transitions maintains diverse experience samples across different phases of strategy evolution, preventing catastrophic forgetting of previously effective tactics when agents adapt to new opponent behaviors.~\cite{catas_forget} The 150-state history window constrains the temporal complexity of strategic reasoning while providing sufficient context for multi-step attack sequences. This parameter directly impacts the computational complexity of state representation and influences the sophistication of temporal strategies both agents can develop. ~\cite{temporal_rl}

\begin{figure*}
    \centering
    \includegraphics[width=0.85\linewidth]{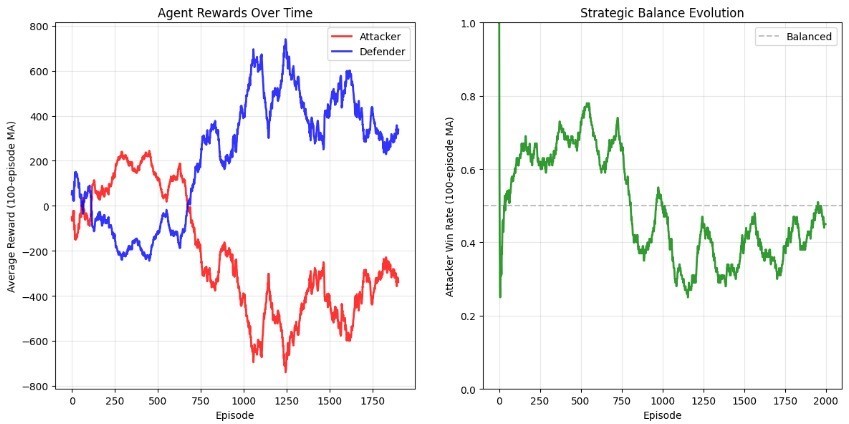}  
    \caption{Learning Performance of Attacker and Defender Agents after adapting them to complex strategies}
    \label{fig:final_learning_rate}
\end{figure*}
 
\subsection{Evaluation}

Figures ~\ref{fig:initial_reward_dynamics}, ~\ref{fig:model1learning_rate}, and ~\ref{fig:final_learning_rate} together illustrate the learning dynamics and strategic balance between the defender and attacker over extended training. While the initial ones span long horizons, the latter emphasize adaptation to richer strategic repertoires.

\textbf{Learning Dynamics:}
In the short-horizon view of Figure~\ref{fig:initial_reward_dynamics}, the defender (blue) quickly achieves large positive returns (peaks near $+60{,}000$) within the first 200 episodes, while the attacker (red) falls sharply to negative values (40,000 to 60,000). This shows that even basic defenses like traps and rate-limiting yield immediate payoff, well before long-term stabilization.

Throughout extended training (Figure~\ref{fig:model1learning_rate}), defensive performance remains consistently positive ($\approx$ +30 to +40 on average), with only brief transients, while the attacker fails to sustain gains. Strategic balance rarely exceeds parity, confirming defender dominance in 50,000 episodes. When agents are exposed to complex strategies (Figure~\ref{fig:final_learning_rate}), defender returns escalate dramatically (+300 to +600), and the attacker win rates stabilize well below parity. Richer defensive options, e.g. adaptive IP blocking, port-specific controls, and traps decisively tilt the equilibrium.

Together, these outcomes reveal a two-phase dynamic. Early training is characterized by steep and volatile divergences (Figure~\ref{fig:initial_reward_dynamics}), with defenders establishing advantage almost immediately. Longer training smooths this volatility (Figure~\ref{fig:model1learning_rate}), while complex strategies (Figure~\ref{fig:final_learning_rate}) drive overwhelming defensive stability.~\cite{volatile_rl} These findings highlight the role of reward shaping, training schedules, and defensive hierarchy in achieving stable convergence in adversarial reinforcement learning.

\section{Results and Discussion}
We summarize the principal findings, aggregated across random seeds and reported as mean values; variances are provided in supplementary logs. Results are organized around ablations, parameter sensitivities, and broader system-level insights.

\subsection{Implementation Notes and Ablations}
We evaluate both \emph{raw} and \emph{shaped} reward variants. Shaping accelerates convergence and reduces variance, but can also alter policy behavior. The false-positive penalty, modeled as a per-request cost when benign traffic is rate-limited, introduces a measurable availability–security tradeoff. Under the baseline configuration ($P_{trap}=60\%$, $T_{min}=300$), defenders converged to conservative strategies emphasizing port-level rate-limiting and selective traps. These policies achieved low false-positive rates while keeping attacker win rates near zero. Reward curves showed oscillations during early training that gradually stabilized with longer horizons. ~\cite{lessfp_rl}

\subsection{Effect of Trap Probability and Exploitation Thresholds}
Trap probability has a strong effect on attacker viability. With $P_{trap}=0\%$, attackers succeeded more frequently but defenders could still suppress them with aggressive rate-limiting, albeit at higher collateral costs. At the default $P_{trap}=60\%$, attacker success declined substantially while defender costs remained moderate. Raising exploitation thresholds ($T_p=400$) further reduced attacker success, though the increased sparsity of terminal rewards slowed attacker learning. Reward shaping mitigated this effect by providing incremental progress signals, but ablations confirm that shaping must be tuned carefully to avoid policy bias.

\subsection{Practical Insights}
Three broader lessons emerge. First, defender observability provides a structural advantage: access to aggregated request histories and IP activity patterns enables robust discrimination of malicious from benign traffic. Second, reward design critically affects attacker learning under sparse feedback; shaping improves stability but risks biasing strategies. Third, training logistics matter: simultaneous learning introduces non-stationarity, while alternating updates, opponent populations, or centralized critics reduce instability. From a systems perspective, traps act as effective honeypot analogs when their detection probability is non-trivial, while selective rate-limiting remains a reliable mechanism for balancing security with service availability.

\section{Conclusion}
We designed and evaluated a compact adversarial reinforcement learning environment that captures key tradeoffs in brute-force exploits and reactive defenses. Our experiments demonstrate how trap mechanisms, exploitation difficulty, reward shaping, and training regimen influence outcomes. The findings emphasize the role of observability and reward engineering in multi-agent learning for security. Future work should explore centralized critics, opponent populations, richer attacker models, and transfer to more realistic network simulators.

\balance

\end{document}